\documentclass[lettersize,journal]{IEEEtran}
\usepackage{amsmath,amsfonts}
\usepackage{algorithmic}
\usepackage{algorithm}
\usepackage{array}
\usepackage[caption=false,font=normalsize,labelfont=sf,textfont=sf]{subfig}
\usepackage{textcomp}
\usepackage{stfloats}
\usepackage{url}
\usepackage{verbatim}
\usepackage{graphicx}
\usepackage{cite}
\usepackage[T1]{fontenc}
\usepackage{multirow}
\usepackage{booktabs}
\usepackage{makecell}
\usepackage{amsmath}
\usepackage{xcolor}    
\usepackage{colortbl}  
\usepackage{tabularx}
\usepackage{lineno,hyperref}
\newcolumntype{Y}{>{\centering\arraybackslash}X}
\begin{document}

\title{NutVLM: A Self-Adaptive Defense Framework against Full-Dimension Attacks for Vision Language Models in Autonomous Driving}

\author{{Xiaoxu Peng, Dong Zhou,~\IEEEmembership{Member,~IEEE,} Jianwen Zhang, Guanghui Sun,~\IEEEmembership{Senior Member,~IEEE,} Anh Tu Ngo, Anupam Chattopadhyay,~\IEEEmembership{Senior Member,~IEEE}}
\thanks{This work was kindly supported by National Natural Science Foundation of China through grant No. 62403162 and Joint Funds of the National Natural Science Foundation of China through grant No. U23A20346. (Corresponding authors: Dong Zhou.)}
\thanks{X. Peng, D. Zhou, and G. Sun are with the Department of Control Science and Engineering, Harbin Institute of Technology, Harbin {\rm 150001}, China (e-mail: xiaoxupeng@stu.hit.edu.cn; dongzhou@hit.edu.cn; guanghuisun@hit.edu.cn).}
\thanks{J. Zhang is with the Department of Electrical Engineering , Harbin Institute of Technology, Harbin {\rm 150001}, China (e-mail: 23b306001@stu.hit.edu.cn).}
\thanks{A. T. Ngo and A. Chattopadhyay are with the College of Computing and Data Science, Nanyang Technological University, Singapore (e-mail: ngoanhtu001@e.ntu.edu.sg; anupam@ntu.edu.sg).}}

\IEEEpubid{0000--0000/00\$00.00~\copyright~2021 IEEE}

\maketitle

\begin{abstract}
Vision Language Models (VLMs) have advanced perception in autonomous driving (AD), but they remain vulnerable to adversarial threats. These risks range from localized physical patches to imperceptible global perturbations. Existing defense methods for VLMs remain limited and often fail to reconcile robustness with clean-sample performance. To bridge these gaps, we propose NutVLM, a comprehensive self-adaptive defense framework designed to secure the entire perception-decision lifecycle. Specifically, we first employ NutNet++ as a sentinel, which is a unified detection-purification mechanism. It identifies benign samples, local patches, and global perturbations through three-way classification. Subsequently, localized threats are purified via efficient grayscale masking, while global perturbations trigger Expert-guided Adversarial Prompt Tuning (EAPT). Instead of the costly parameter updates of full-model fine-tuning, EAPT generates "corrective driving prompts" via gradient-based latent optimization and discrete projection. These prompts refocus the VLM’s attention without requiring exhaustive full-model retraining. Evaluated on the Dolphins benchmark, our NutVLM yields a 4.89\% improvement in overall metrics (e.g., Accuracy, Language Score, and GPT Score). These results validate NutVLM as a scalable security solution for intelligent transportation. Our code is available at \url{https://github.com/PXX/NutVLM}. 
\end{abstract}


\section{Introduction}
\IEEEPARstart{T}{he} rapid evolution of VLMs has driven a paradigm shift in AD, advancing on-board intelligence from basic perceptual recognition to complex semantic reasoning~\cite{zhou2024vision, ngo2024systematic, wang2024toward}. By aligning visual data with linguistic instructions, these models enable vehicles to interpret diverse traffic scenarios and handle long-tail driving situations~\cite{tang2025video}. However, this multimodal fusion broadens the attack surface, exposing AD systems to diverse adversarial threats~\cite{he2025artificial}.

\begin{figure}[t]
	\centering	
	{\includegraphics[scale=0.4]{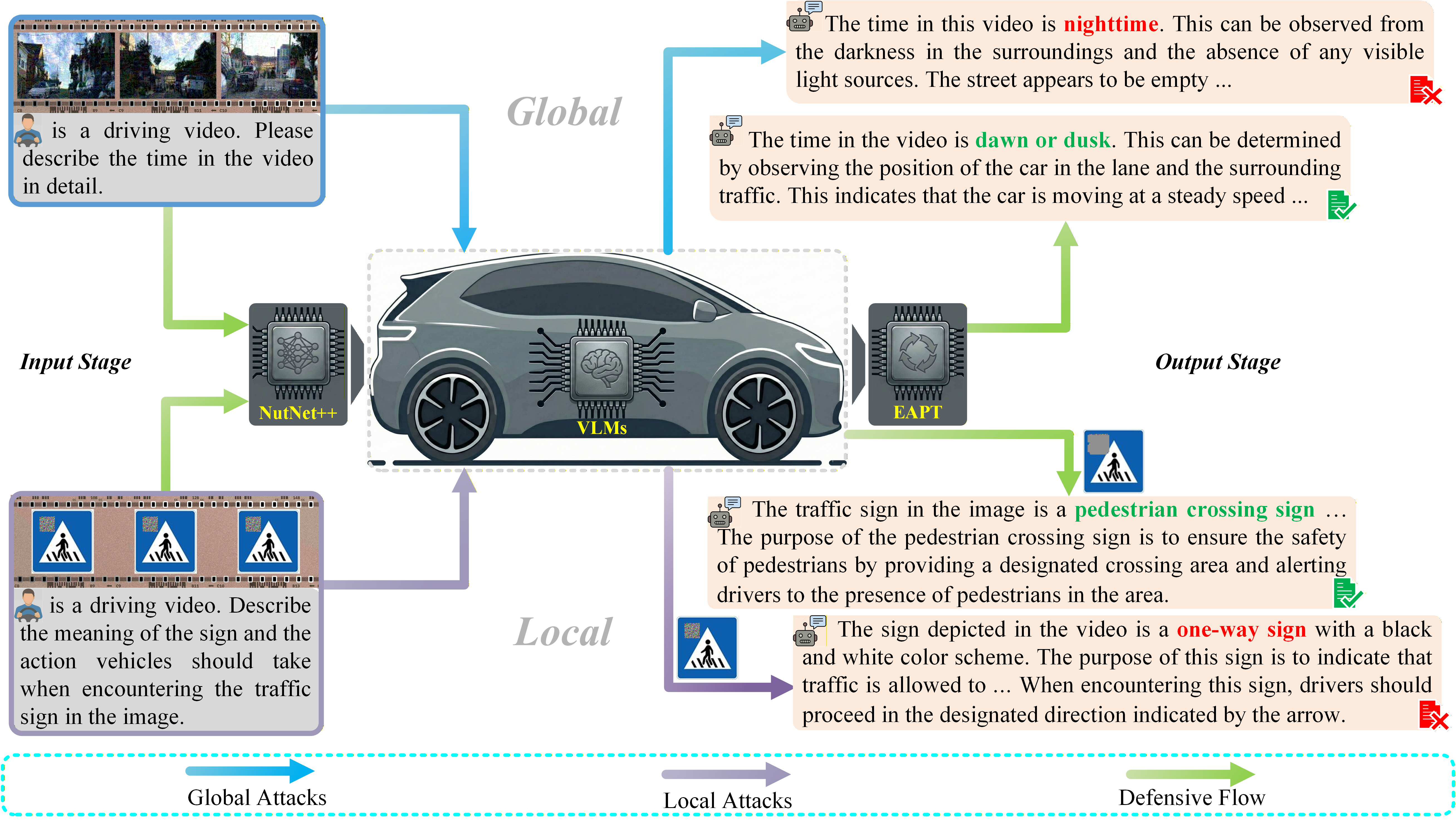}}
	\captionsetup{font={footnotesize}}
	\caption{Intuitive example of verifying the security of NutVLM in adversarial scenarios. Standard VLMs can be misled by adversarial examples and generate hazardous commands (marked in red), while NutVLM detects and purifies the input to yield safe navigation instructions (marked in green).}
	\label{fig_1}
\end{figure}

\IEEEpubidadjcol

Earlier research primarily targeted individual perception modules or small-scale AD networks. For instance, attackers have successfully targeted traffic sign recognition systems via physical patches~\cite{wang2025unified} and disrupted video object segmentation (VOS) through hard region discovery~\cite{li2023adversarial}, causing navigation failures. In more complex Bird's-Eye-View (BEV) perception, AdvRoad~\cite{wang2025invisible} demonstrates that road-style adversarial posters can induce false-positive obstacles, thereby misleading the vehicle's 3D spatial understanding. These methods suggest that AD attacks are predominantly divided into local attacks~\cite{ran2023cross} utilizing physical patches on semantic targets and global attacks~\cite{dong2025feature} disrupting cognitive alignment through frame-wide noise. 

As VLMs become increasingly integrated into the AD pipeline, the landscape of adversarial threats has flourished, shifting from pixel-level perturbations to the disruption of multimodal reasoning integrity. Depending on the degree of epistemic access granted to the target model, adversarial attacks are dichotomized into white-box~\cite{mkadry2017towards} and black-box~\cite{chen2017zoo} settings. Within the white-box domain, BadVLMDriver~\cite{ni2024physical} shows that physical backdoor triggers can be embedded into models to hijack outputs. Furthermore, ADvLM~\cite{zhang2024visual} exploits textual instructions and temporal dependencies to degrade driving performance while demonstrating black-box transferability. Notably, CADA~\cite{wang2025black} recently employs black-box decision chain disruption and risk scenario induction to force catastrophic commands by interrupting the logical flow of multimodal reasoning.

To mitigate these threats, various defense strategies have been proposed to bolster the robustness of AD VLMs. Adversarial training (AT)~\cite{mkadry2017towards} based methods remain impractical for large-scale VLMs due to prohibitive computational costs and the degradation of clean-sample performance. Consequently, research has pivoted toward inference-time detection and purification. For instance, LightPure~\cite{khalili2024lightpure} employs a Diffusion-GAN hybrid for rapid image purification on mobile hardware, while LVLM-FDA~\cite{chen2025lvlm} leverages internal representations across attention heads to identify malicious attempts with minimal computational overhead. Despite these advances, current defense methods struggle to reconcile real-time efficiency with robustness against full-dimension threats (encompassing both local and global attacks). They also lack validation on state-of-the-art (SOTA) VLM-based AD platforms such as Dolphins~\cite{ma2024dolphins}, DriveLM~\cite{sima2024drivelm}, and LMDrive~\cite{shao2024lmdrive}.

To address these limitations, we propose NutVLM, a self-adaptive online defense framework compatible with these platforms. This framework incorporates NutNet++ (an enhanced variant of NutNet~\cite{lin2024don}) at the input stage for unified threat detection and immediate physical attack purification. It further triggers EAPT for real-time instruction correction to resolve perceptual ambiguities in complex scenarios. Remarkably, NutVLM addresses the robustness-accuracy trade-off by maintaining resilience against full-dimension threats while outperforming baseline accuracy under adversarial conditions. The framework is validated across two general datasets and one AD-specific Dolphins benchmark. To ensure architecture-agnostic generalizability, we evaluate the system against 7 full-dimension threats using three additional VLM backbones: InstructBlip~\cite{dai2023instructblip}, LLaVA~\cite{liu2023visual}, and MiniGPT-v4~\cite{zhu2023minigpt}. Within the CADA benchmark, NutVLM is compared against 6 representative defense baselines, achieving a 73.94\% F1-score while maintaining high inference speed. Average defensive improvements reach 1.06\% for global attacks and 3.83\% for local attacks, demonstrating superior and universal efficacy across diverse adversarial environments.

In summary, the specific contributions of our study are as follows:

\begin{itemize}
	\setlength{\leftmargin}{0pt}
	\item
	We propose NutVLM as a comprehensive self-adaptive online defense framework for AD VLMs, which is capable of mitigating full-dimension adversarial threats.
	\item
	We introduce the NutNet++ sentinel for unified attack detection and purification, employing pixel-level grayscale masking to neutralize physical threats.
	\item 
	We propose EAPT for real-time, parameter-free instruction correction via gradient-based latent optimization and discrete projection, whose superior defense generalization and effectiveness are validated through extensive experiments.
\end{itemize}

The remainder of this article is structured as follows: Section 2 reviews related work on adversarial attacks and defense strategies in VLMs. Section 3 elaborates on the methodology of the NutVLM framework. Section 4 presents the experimental results and ablation studies, and Section 5 concludes the paper.

\section{Related Work}
This section traces the evolution of VLM attacks from unimodal interference to cross-modal reasoning disruptions. Then we further evaluate current defense mechanisms while highlighting the computational bottlenecks that limit their practical deployment.

\subsection{Adversarial Attacks on VLMs}

The logic of attacking VLMs initially inherited traditional CNN-based strategies targeting image encoders, such as PGD~\cite{mkadry2017towards} and AutoAttack~\cite{croce2020reliable}. However, as downstream tasks grew more complex, research pivoted towards upstream attacks~\cite{peng2025rpf} on pre-trained VLM backbones. For example, AdvCLIP~\cite{zhou2023advclip} relies on adversarial transferability through surrogate models to mislead target VLMs without gradient access. AnyAttack~\cite{zhang2024anyattack} employs self-supervised learning to compromise diverse architectures. 

This focus subsequently expanded to text encoders, enabling multimodal co-optimization. White-box methods like Co-Attack~\cite{zhang2022towards} concurrently perturb image and text modalities to maximize feature mismatch. Subsequently, SGA~\cite{lu2023set} extended this paradigm to black-box settings. Meanwhile, localized threats such as PiE~\cite{kong2024patch} have also demonstrated that confined patterns could hijack the attention of pre-trained models. 

Integrating VLMs into AD systems escalates threats from simple perceptual errors to critical decision-chain disruptions. Zhang et al.~\cite{zhang2024visual} successfully altered predicted waypoints by exploiting textual instructions and temporal dependencies, demonstrating significant black-box transferability of ADvLM. In the physical domain, PhysPatch~\cite{guo2025physpatch} further reveals that realizable patches placed on road signs can consistently mislead multimodal agents across varying viewing angles. Furthermore, CADA~\cite{wang2025black} utilizes risky scene induction to dismantle the causal reasoning required for navigation, encompassing both local and global adversarial threats. These evolving attacks underscore the urgent need for more effective defense methods.

\subsection{Adversarial Defenses for VLMs}
Parameter-update VLM defenses primarily originate from standard AT~\cite{mkadry2017towards}. TeCoA~\cite{mao2022understanding}, FARE~\cite{schlarmann2024robust}, and MMCoA~\cite{zhou2024revisiting} harden backbones via adversarial fine-tuning or adversarial contrastive tuning but still face high computational costs. To reduce overhead, parameter-efficient methods including AdvPT~\cite{zhang2024adversarial} and APT~\cite{li2024one} bolster robustness by optimizing textual prompts without modifying model weights. While FAP~\cite{zhou2024few} further refines cross-modal consistency, these training-time approaches lack the agility required for unforeseen threats. TAPT~\cite{wang2025tapt} introduces test-time prompt adaptation to provide inference-stage flexibility, yet it remains vulnerable to localized adversarial patches.

Inference-time mechanisms prioritize identifying and neutralizing threats without altering model parameters. Detection frameworks such as PIP~\cite{zhang2024pip} and DPS~\cite{zhou2024defending} utilize irrelevant probe questions or consistency checks to identify attacks, but their reliance on binary rejection often results in service interruptions. MirrorCheck~\cite{fares2024mirrorcheck} utilizes Text-to-Image (T2I) regeneration loops to verify semantic consistency. However, the resulting generative latency is incompatible with the real-time constraints of AD.

General VLM defenses rarely account for AD-specific requirements, particularly the need for semantic-logic alignment and resilience against full-dimension threats. LightPure~\cite{khalili2024lightpure} enables rapid image purification on mobile hardware; however, it overlooks the linguistic ambiguities caused by adversarial manipulation. LVLM-FDA~\cite{chen2025lvlm} leverages internal representations across attention heads to efficiently identify malicious features during inference, while AutoTrust~\cite{xing2024autotrust} quantifies DriveVLM~\cite{tian2024drivevlm} vulnerabilities across five trust pillars. These remain diagnostic or binary detection tools rather than active, self-adaptive defenses.

Addressing these unmet needs, we propose NutVLM, a holistic online defense framework designed to balance multi-dimensional robustness with real-time inference efficiency. It secures the entire perception-decision lifecycle without requiring offline retraining or high-latency generative loops, ensuring safe and continuous operation on SOTA AD platforms.
\begin{figure*}[t]
	\centering	
	{\includegraphics[scale=0.58]{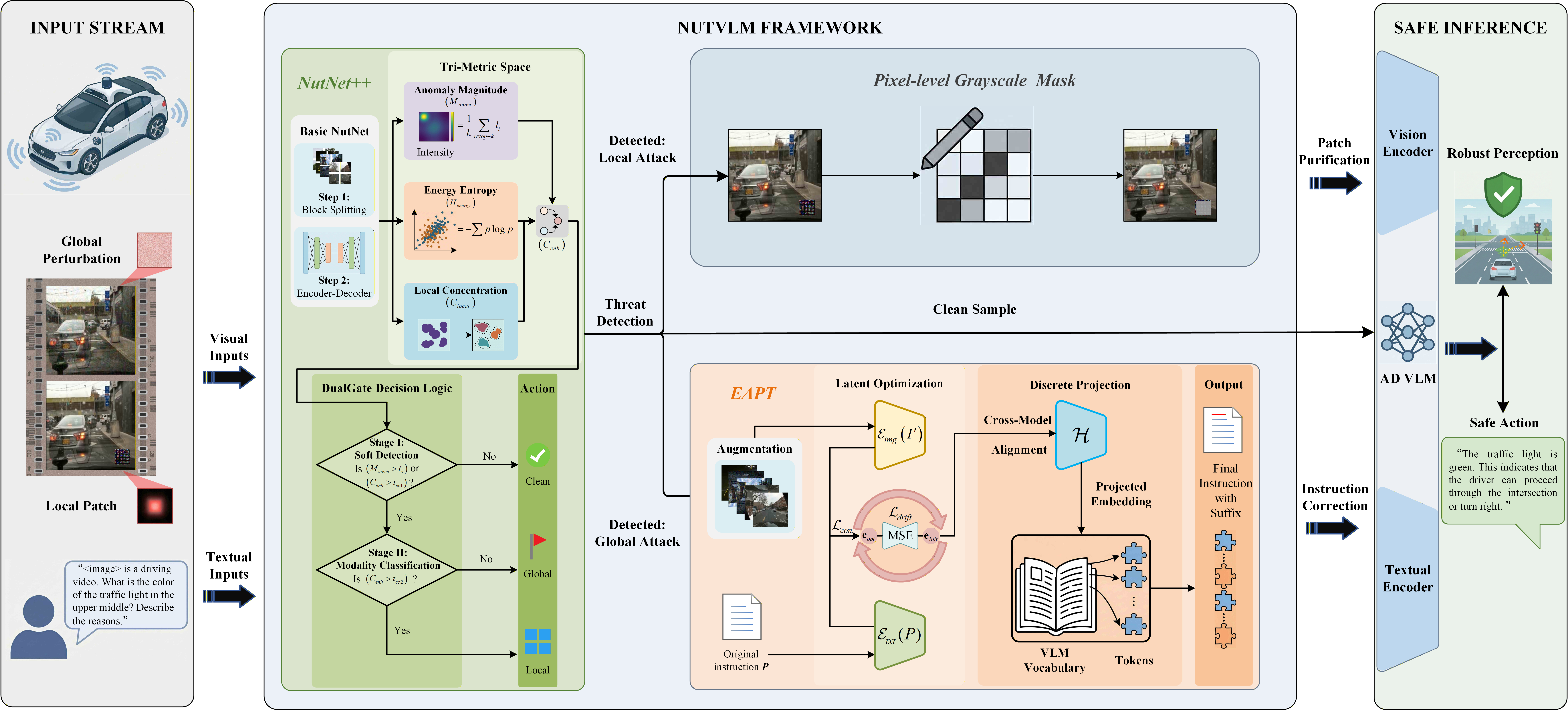}}
	\captionsetup{font={footnotesize}}
	\caption{Overview of the NutVLM defense framework. The architecture integrates NutNet++ as a unified detection module to identify local patches and global perturbations. Depending on the detected threat, the framework executes either pixel-level grayscale masking for patch purification or EAPT for instruction correction. This dual-branch approach ensures robust perception and safe inference for AD VLMs.}
	\label{fig_2}
\end{figure*}

\section{NutVLM Framework Overview}
This section provides a comprehensive overview of the proposed NutVLM framework, as illustrated in Figure~\ref{fig_2}. We begin by establishing the preliminaries for AD VLMs and defining the adversarial attack settings. Subsequently, we detail the architectural design and mathematical formulation of NutVLM. The framework is centered around two core components, namely the NutNet++ detection module for threat identification and the EAPT module for real-time instruction correction and defense.

\subsection{Preliminaries and Adversarial Attacks}

\subsubsection{AD VLM Preliminaries}
In the context of AD, a VLM, denoted as $\mathcal{F}_{\theta}$, serves as a centralized brain. Formally, given visual inputs $I \in \mathbb{R}^{T \times H \times W \times C}$ and a clean textual prompt $P$, the model predicts the driving response $Y$:

\begin{equation}
	\label{eq:1}
	Y = \mathcal{F}_{\theta}(I, P),
\end{equation}

where $\theta$ represents the model parameters. The dimensions $T, H, W$, and $C$ denote the temporal sequence length, image height, width, and channels, respectively.

\subsubsection{Adversarial Attacks}

We focus on visual adversarial attacks, where the adversary injects perturbations solely into the image modality while the text prompt $P$ remains clean. Specifically, we consider two standard categories of threats prevalent in AD:

\begin{itemize}
	\item \textbf{Global Attacks:} The adversary adds imperceptible noise $\delta_{global}$ across the entire image. These perturbations can be guided by either visual gradients or textual semantics to disrupt cross-modal alignment. The adversarial example is defined as:
	\begin{equation}
		\label{eq:2}
		I' = I + \delta_{global}, \quad \text{s.t.} \quad \|\delta_{global}\|_p \le \epsilon,
	\end{equation}
	
	where $\|\cdot\|_p$ denotes the $L_p$ norm (typically $L_\infty$ or $L_2$) and $\epsilon$ represents the perturbation budget.
	
	\item \textbf{Local Attacks:} The adversary applies a visible, high-intensity patch to a specific region defined by a binary mask $M \in \{0, 1\}^{H \times W}$. The physical-world threat is formulated as:
	\begin{equation}
		\label{eq:3}
		I' = (1 - M) \odot I + M \odot \delta_{patch},
	\end{equation}
	
	where $\odot$ denotes element-wise multiplication, and $\delta_{patch}$ represents the adversarial content within the masked region.
	
\end{itemize}

In both cases, the adversarial example $I'$ is optimized to maximize the prediction error or target a hazardous semantic output.

\subsection{NutVLM}
\subsubsection{Defense Objective}
To counteract these visual threats, NutVLM aims to learn a robust mapping by integrating visual-level purification and instruction-level prompt correction. Serving as the primary detector, NutNet++ distinguishes global perturbations, local patches, and clean samples. For identified local attacks, NutNet++ applies a physical purification operation $\mathcal{P}_{mask}(\cdot)$ to neutralize the attack patch, yielding a purified visual input $I^* = \mathcal{P}_{mask}(I')$ while preserving the global context for other categories.

Subsequently, to ensure reliable adaptation and balance efficiency with robustness, we deploy EAPT, denoted as $\mathcal{T}(\cdot)$. Guided by the detection logic, EAPT generates a robust prompt $P^* = \mathcal{T}(P)$ when global adversarial threats are confirmed. This mechanism enables the model to ignore residual imperceptible noise by calibrating the input instructions before inference, thereby avoiding unnecessary linguistic variance for clean or successfully purified samples.

Consequently, the unified defense objective is formalized as minimizing the divergence between the model prediction and the ground truth, as shown in the following formulation.
\begin{equation}
	\label{eq:4}
	\min_{\mathcal{P}_{mask}, \mathcal{T}} \mathcal{L}\left(\mathcal{F}_{\theta}\left(I^*, P^*\right), Y_{gt}\right).
\end{equation}
Here, the synergy of $\mathcal{P}_{mask}$ and $\mathcal{T}$ provides a unified defense shield safeguarding both the visual and textual input modalities to ensure end-to-end functional safety.

\begin{figure*}[t]
	\centering
	\begin{minipage}{0.24\textwidth}
		\centering
		\includegraphics[width=\textwidth]{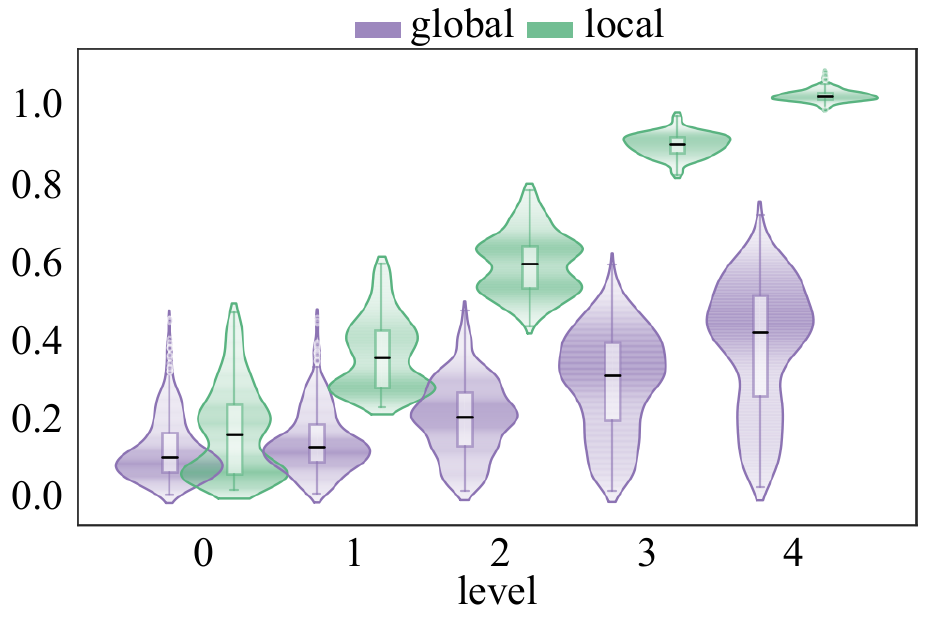}
		\makebox[\textwidth]{\footnotesize (\textbf{a}) $M_{anom}$}
		\label{fig_3_1}
	\end{minipage}
	\hfill
	\begin{minipage}{0.24\textwidth}
		\centering
		\includegraphics[width=\textwidth]{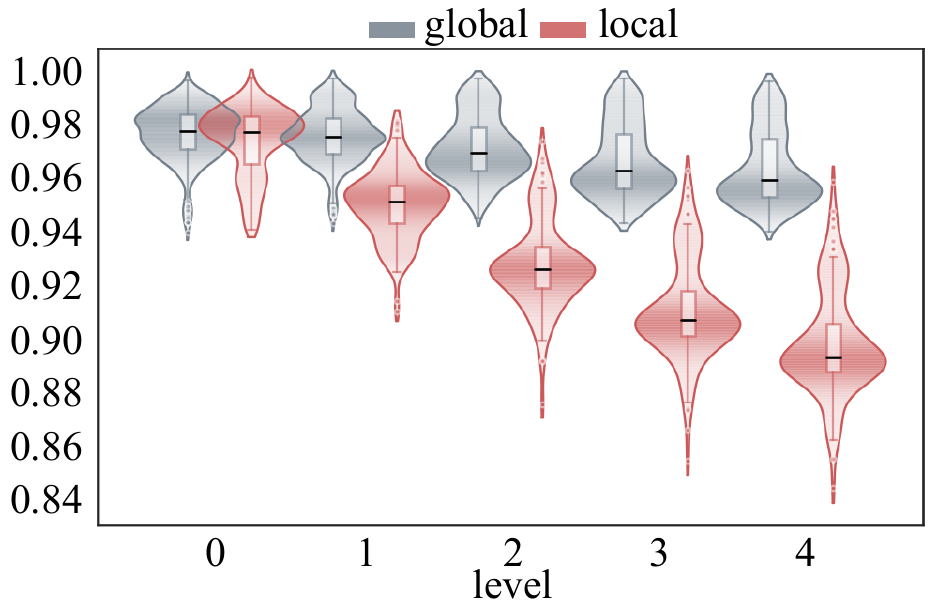}
		\makebox[\textwidth]{\footnotesize (\textbf{b}) $H_{energy}$}
		\label{fig_3_2}
	\end{minipage}
	\hfill
	\begin{minipage}{0.24\textwidth}
		\centering
		\includegraphics[width=\textwidth]{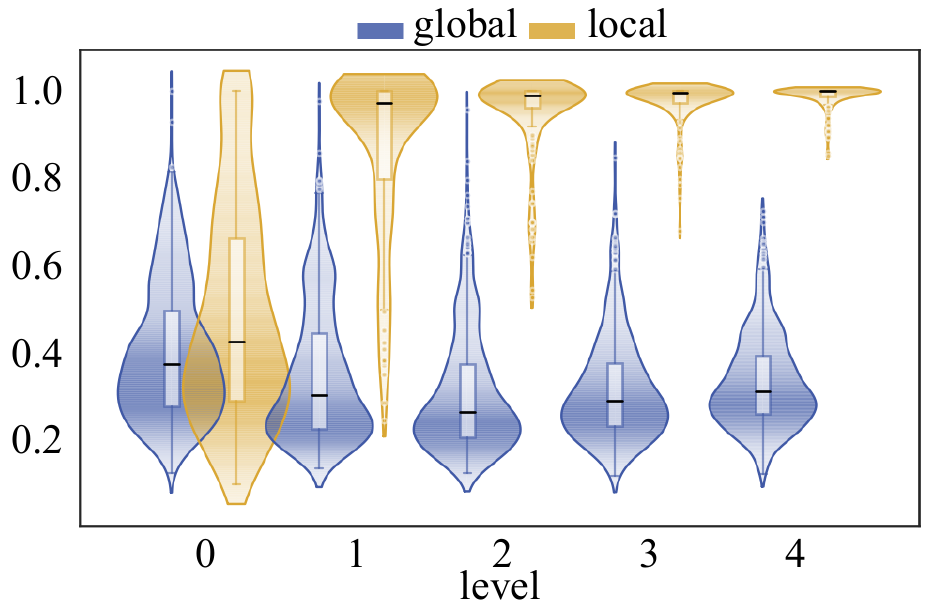}
		\makebox[\textwidth]{\footnotesize (\textbf{c}) $C_{local}$}
		\label{fig_3_3}
	\end{minipage}
	\hfill
	\begin{minipage}{0.24\textwidth}
		\centering
		\includegraphics[width=\textwidth]{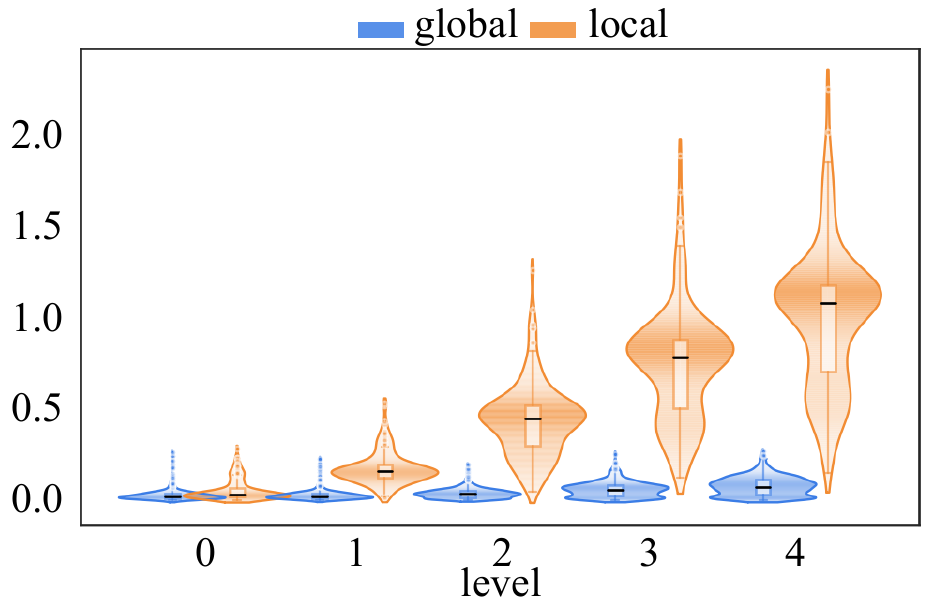}
		\makebox[\textwidth]{\footnotesize (\textbf{d}) $C_{enh}$}
		\label{fig_3_4}
	\end{minipage}
	\caption{Statistical distributions of key defense metrics across varying CADA attack intensities (both global and local adversarial scenarios). (a) Anomaly Magnitude $M_{anom}$, (b) Energy Entropy $H_{energy}$, (c) Local Concentration $C_{local}$, and (d) the Enhanced Concentration $C_{enh}$.}
	\label{fig_3}
\end{figure*}

\subsubsection{NutNet++}

While the original NutNet~\cite{lin2024don} excels in reconstruction-based out-of-distribution (OOD) detection against physical patches, it treats all anomalies uniformly as local attacks, leaving it vulnerable to stealthier threats. Existing defenses often overlook the intrinsic energy distribution disparities between global perturbations and local patches. To mitigate this limitation, we propose NutNet++, which advances the binary detection to a three-way sensing mechanism covering clean samples, global attacks, and local attacks. As illustrated in Fig.~\ref{fig_2}, this module dynamically orchestrates the defense branches by adaptively switching between physical purification for local attacks and instruction-level correction for global attacks.

\textbf{Metric Formulation and Feature Analysis.} To effectively distinguish clean samples from adversarial threats, we propose three core metrics derived from reconstruction error patterns. Specifically, $M_{anom}$ identifies the presence of anomalies, while $H_{energy}$ and $C_{local}$ collectively differentiate between diffused global noise and concentrated local patches based on their distinct spatial signatures.

\begin{itemize}
	\item \textit{Anomaly Magnitude} ($M_{anom}$) quantifies the overall intensity of the reconstruction errors. To ensure robustness across AD video streams, we adopt a Conditional Value-at-Risk (CVaR) strategy~\cite{rockafellar2000optimization}.
	\begin{equation}
		\label{eq:5}
		M_{anom} = \mathbb{E}\left[\ell \mid \ell \ge \text{VaR}_\alpha(\ell)\right],
	\end{equation}
	where $\ell$ represents the random variable of block-wise reconstruction losses and $\mathbb{E}[\cdot]$ denotes the expectation operator. The term $\text{VaR}_\alpha(\ell)$ signifies the Value-at-Risk at confidence level $\alpha$ (empirically set to $0.95$). This approach effectively mitigates temporal jitter by focusing on the tail distribution of losses. For single-frame inputs, anomaly magnitude value simplifies to the standard reconstruction loss.
	
	\item \textit{Energy Entropy ($H_{energy}$)} characterizes the spatial dispersion of residual energy. Global attacks typically exhibit high entropy due to diffused disturbances, while local attacks manifest concentrated energy distributions. This metric utilizes the Shannon entropy~\cite{shannon1948mathematical} of the normalized loss vector $e$.
	\begin{equation}
		\label{eq:6}
		H_{energy} = - \sum_{b \in \mathcal{B}} (e_b \log e_b), \quad \text{with } e_b = \frac{\ell_b}{\sum_{j \in \mathcal{B}} \ell_j},
	\end{equation}
	where $\ell_b$ denotes the reconstruction loss of block $b \in \mathcal{B}$. Elevated energy entropy values indicate uniform, noise-like disturbances characteristic of global attacks.
	
	\item \textit{Local Concentration ($C_{local}$)} acts as the primary discriminator between local and global attacks by quantifying the spatial aggregation of perturbations. We calculate the weighted share of the largest connected component to account for attack intensity.
	\begin{equation}
		\label{eq:7}
		C_{local} = \frac{\max_{k} \left( \sum_{b \in \mathcal{S}_k} \ell_b \right)}{\sum_{b \in \mathcal{B}} \ell_b},
	\end{equation}
	where $\mathcal{B}$ denotes the set of all blocks within the image and $\mathcal{S}_k$ represents the set of blocks belonging to the $k$-th connected component in the thresholded error map. High local concentration values reveal that reconstruction errors are concentrated within a single contiguous region which identifies the presence of local attacks.
\end{itemize}

The empirical analysis visualized in Fig.~\ref{fig_3} validates the effectiveness of these metrics. As shown in the data distributions, local attacks are characterized by high spatial concentration and low energy dispersion, whereas global attacks exhibit significant magnitude rises with diffused energy patterns. These distinct statistical signatures provide a reliable basis for the subsequent classification logic.

\textbf{DualGate Decision Logic.} We present a DualGate strategy based on the observed statistical distributions, which utilizes a two-stage discrimination logic to ensure high-precision threat identification.

\textit{Stage I: Detection with Soft-Recall.}
The initial stage distinguishes clean and adversarial examples. To address the limitations of rigid thresholds, we introduce an enhanced concentration indicator $C_{enh}$.
\begin{equation}
	\label{eq:8}
	C_{enh} = C_{local} \cdot (1 - \tilde{H}_{energy})^\beta,
\end{equation}
here $\tilde{H}_{energy}$ is the normalized entropy and $\beta=0.8$ is a concentration factor. As demonstrated in Fig.~\ref{fig_3} (d), $C_{enh}$ effectively amplifies the structural features of local attacks while suppressing noise in clean samples. The Soft-Recall logic is then formulated as the following piecewise function.
\begin{equation}
	\label{eq:9}
	Attacked = \begin{cases} 
		1, & M_{anom} > t_s \\ 
		1, & M_{anom} \le t_s \wedge C_{enh} > t_{cc1} \\ 
		0, & \text{otherwise} 
	\end{cases}
\end{equation}
where $t_{cc1}$ is the sensitivity threshold for concentration. This ensures that samples with low magnitude but high structural anomaly are recalled, reducing false negatives.

\textit{Stage II: Modality Classification.}
Upon confirming an attack, the second stage performs fine-grained classification. Since local attacks exhibit strong signals on $C_{enh}$, we use a specific threshold $t_{cc2}$ to separate them from global attacks.
\begin{equation}
	\label{eq:10}
	Class = \begin{cases} 
		\text{Local attack}, & \text{if } C_{enh} > t_{cc2} \\ 
		\text{Global attack}, & \text{otherwise}.
	\end{cases}
\end{equation}

This hierarchical pipeline ensures that each identified threat is precisely categorized to facilitate the activation of targeted defense mechanisms within NutVLM.

\subsubsection{Expert-guided Adversarial Prompt Tuning}
For confirmed global attacks, NutVLM invokes EAPT to perform semantic-level rectification. EAPT leverages a frozen CLIP model as an experienced Expert to provide robust cross-modal anchors. By generating a corrective suffix $S$, this module calibrates the misalignment between the perturbed visual input $I'$ and the original instruction $P$. This process forces the AD VLMs to re-attend to critical task-relevant features and neutralize adversarial hallucinations.

\textbf{Semantic Verification Gate.} The activation of EAPT is controlled by a verification gate to ensure system efficiency. This process utilizes a Semantic Verification Score $V_{sem}$ to quantify the alignment between the visual features and the linguistic intent. Since the Expert’s image encoder $\mathcal{E}_{img}$ and text encoder $\mathcal{E}_{txt}$ share a unified latent space, the consistency is measured via the following score.
\begin{equation}
	\label{eq:11}
	V_{sem} = \text{cos}\left(\mathcal{E}_{img}(I'), \mathcal{E}_{txt}(P)\right),
\end{equation}
where $\text{cos}(\cdot, \cdot)$ denotes the cosine similarity. A low $V_{sem}$ indicates that adversarial perturbations have successfully compromised the semantic link, thereby triggering the downstream optimization only when $V_{sem} < \tau_{sem}$. Otherwise, the original prompt $P$ is passed directly to ensure system efficiency. In our implementation, we empirically set $\tau_{sem} = 0.2$ to filter out environmental noise while capturing genuine semantic attacks.

\textbf{Latent Optimization with Drift Control.} Once triggered, EAPT executes a forward optimization to find an optimal latent embedding $\mathbf{e}_{opt}$. This vector is initialized as $\mathbf{e}_{init} = \mathcal{E}_{\text{txt}}(P)$, representing the initial embedding of the original instruction $P$ within the Expert's latent space. Subsequently, $\mathbf{e}_{opt}$ is refined iteratively for $K$ steps to restore image-text consistency while remaining anchored to the original driving task. EAPT minimizes a joint objective function.
\begin{equation}
	\label{eq:12}
	\mathcal{L}_{EAPT} = \mathcal{L}_{con} + \lambda \mathcal{L}_{drift},
\end{equation}
the consistency loss $\mathcal{L}_{con}$ encourages the optimized embedding to capture the true visual context verified by the Expert and $\lambda$ is a balancing coefficient set to 0.1.
\begin{equation} 
	\label{eq:13} 
	\mathcal{L}_{con} = \mathbb{E}_{\tilde{I} \sim \textit{T}(I')}\left[1 - \text{cos}\left(\mathcal{E}_{\text{img}}(\tilde{I}), \mathbf{e}_{opt}\right)\right],
\end{equation}
here $\mathbb{E}[\cdot]$ denotes the expectation over stochastic augmentations $\textit{T}(\cdot)$. Simultaneously, the drift loss $\mathcal{L}_{drift}$ prevents the optimization from deviating from the Expert’s original intent by penalizing the distance from the initial instruction embedding $\mathbf{e}_{init}$.
\begin{equation}
	\label{eq:14}
	\mathcal{L}_{drift} = ||\mathbf{e}_{opt} - \mathbf{e}_{init}||^2.
\end{equation}

\textbf{Suffix Projection and Attention Reinforcement.} To materialize the continuous embedding $\mathbf{e}_{opt}$ into a format recognizable by the AD VLM, EAPT employs a pre-aligned projector $\mathcal{H}$ and performs a nearest-neighbor search (NNSearch) within the VLM vocabulary $\mathbf{E}_{v}$.
\begin{equation}
	\label{eq:15}
	\mathbf{e}^*_{opt} = \mathcal{H}(\mathbf{e}_{opt}) = \mathbf{W} \cdot \mathbf{e}_{opt} + \mathbf{b},
\end{equation}
\begin{equation}
	\label{eq:16}
	t_k = \arg\max_{t \in \mathcal{V}} \left[ \text{cos}(\mathbf{e}^*_{opt}, \mathbf{E}_{v}[t]) \right].
\end{equation}

The resulting tokens $S = \{t_1, t_2, \dots, t_k\}$ act as high-weight semantic anchors. An instantiated example of the generated suffix $S$ is shown as follows.
\begin{equation}
	\label{eq:17}
	S = [\underbrace{\texttt{pleasedel}, \texttt{remncar}}_{\text{semantic negation}}, \underbrace{\texttt{thisvideo}, \texttt{del}}_{\text{context anchor}}, \underbrace{\texttt{!}, \texttt{!}, \texttt{!}}_{\text{attention gain}}].
\end{equation}

As illustrated in the output of NutVLM in Fig.~\ref{fig_4}(a), the final robust prompt $P^* = P \oplus S$ reinforces the attention mechanism, ensuring the AD VLM remains focused on safe driving actions despite the presence of global adversarial noise. 

By synergizing NutNet++ and EAPT, NutVLM achieves a comprehensive defense across both visual and textual modalities. This integrated procedure is formally outlined in Algorithm~\ref{alg:nutvlm}.

\begin{algorithm}[t]
	\small 
	\caption{NutVLM Defense Pipeline}
	\label{alg:nutvlm}
	
	\begin{algorithmic}[1]
		\renewcommand{\algorithmicrequire}{\textbf{Input:}}
		\renewcommand{\algorithmicensure}{\textbf{Output:}}
		
		\REQUIRE Perturbed frame $I'$, User prompt $P$.
		\REQUIRE AD VLM $\mathcal{F}_{\theta}$, CLIP Encoders $\mathcal{E}_{\text{img}} / \mathcal{E}_{\text{txt}}$, Vocabulary $\mathbf{E}_{v}$.
		\REQUIRE Hyperparams $\{t_s, t_{cc1}, t_{cc2}, \tau_{sem}, K, \eta\}$.
		
		\STATE Compute $M_{anom}, C_{enh}$ via Eq.~\eqref{eq:5}, \eqref{eq:8}.
		
		\IF{$(M_{anom} > t_s) \lor (C_{enh} > t_{cc1})$}
		\IF{$C_{enh} > t_{cc2}$} 
		\STATE $I^* \leftarrow \operatorname{Mask}(I')$  
		\ELSE
		\STATE Compute $V_{sem}$ via Eq.~\eqref{eq:11}.
		\IF{$V_{sem} < \tau_{sem}$}
		\STATE Initialize embedding $\mathbf{e}_{init} \leftarrow \mathcal{E}_{\text{txt}}(P)$.
		\FOR{$k = 1 \text{ to } K$} 
		\STATE Sample augmented view $\tilde{I} \sim \textit{T}(I')$.
		\STATE Compute $\mathcal{L}_{EAPT}$ via Eq.~\eqref{eq:12}.
		\STATE Update $\mathbf{e}_{opt} \leftarrow \mathbf{e}_{opt} - \eta \nabla_{\mathbf{e}_{opt}} \mathcal{L}_{EAPT}$.
		\ENDFOR
		\STATE $S \leftarrow \operatorname{NNSearch}(\mathcal{H}(\mathbf{e}_{opt}), \mathbf{E}_{v})$ via Eq.~\eqref{eq:16}.
		\STATE $P^* \leftarrow P \oplus S$
		\ENDIF
		\ENDIF
		\ENDIF
		
		\STATE \textbf{Output:} $Y \leftarrow \mathcal{F}_{\theta}(I^*, P^*)$
	\end{algorithmic}
\end{algorithm}

\section{Experimental Results and Analysis}
In this section, we conduct a structured set of experiments to systematically verify the effectiveness of NutVLM in securing SOTA VLM-based AD platforms. The experimental design assesses the robustness and generalization of the framework against heterogeneous threats. Furthermore, extensive ablation studies are conducted to disentangle and quantify the individual contributions of attack detection, visual purification, and instruction correction to the overall defense performance.

\subsection{Experimental Setup}

\subsubsection{Implementation Details}
We implemented NutVLM using PyTorch on a HPC server with an Intel Xeon Gold 6132 CPU and one RTX 4090Ti, utilizing CUDA 11.8 and FP16 precision. Results are averaged over five trials with fixed seeds to ensure statistical reproducibility. We employed CLIP (ViT-B/16)~\cite{radford2021learning} as the experienced Expert. Operational hyperparameters involve thresholds set at $t_s=0.2$, $t_{cc1}=0.03$, $t_{cc2}=0.02$, and $\tau_{sem}=0.2$ while the optimization sequence utilizes $K=3$ iterations with a EAPT learning rate of $\eta=5 \times 10^{-3}$. Regarding the comparative baselines, we strictly followed the default configurations from their original papers unless otherwise noted.

\subsubsection{Target Models and Datasets}
To assess the universality and transferability of NutVLM, we evaluate it on Visual Question Answering (VQA) tasks across four representative VLM backbones adapted for AD. Including the domain-specific Dolphins (DP)~\cite{ma2024dolphins} (built on OpenFlamingo~\cite{awadalla2023openflamingo}) and general-purpose InstructBlip (IB)~\cite{dai2023instructblip}, LLaVA (LV)~\cite{liu2023visual}, and MiniGPT-v4 (MG)~\cite{zhu2023minigpt}. 

For the AD-specific dataset, we uniformly utilize the CADA video benchmark~\cite{wang2025black} for evaluations, which consists of two primary subsets.
\begin{itemize}
	\item \textbf{Scene-CADA} incorporates six distinct driving scenarios encompassing Weather (Weath.), Traffic\_light (Traf.), Time\_of\_day (Time),  Scene, Open\_Voc\_Object (Obj.), and Detailed\_description (Desc.).
	\item \textbf{Obj-CADA} focuses on three critical traffic signs including Mandatory\_sign, Prohibition\_sign, and Warning\_sign.
\end{itemize}

Both subsets feature increasing attack intensities ranging from Level 0 (clean) to Level 4 (maximum perturbation). Notably, Level 0 serves as the clean baseline used for implementing other comparative attacks and defenses.

For the general dataset, we incorporate two complementary datasets to evaluate NutNet++ against diverse attack patterns.

\begin{itemize} 
	\item \textbf{APRICOT}~\cite{braunegg2020apricot} comprises over 1,000 annotated real-world images of physical adversarial patches. These samples, targeting COCO objects across diverse perspectives, serve as a rigorous benchmark for localized physical anomalies. 
	\item \textbf{DAmageNet}~\cite{chen2020universal} provides a large-scale library of 50,000 ImageNet-derived samples. Leveraging attention-distraction mechanisms, it simulates highly transferable and aggressive global digital perturbations. 
\end{itemize}

\subsubsection{Compared Attacks}
We constructed a comprehensive threat landscape comprising both standard and adaptive adversarial attacks to test the defense boundary of NutVLM:
\begin{itemize}
	\item \textbf{Traditional Attacks}. To assess generalization against conventional threats, we incorporate established baselines including white-box methods (FGSM~\cite{goodfellow2014explaining}, PGD~\cite{mkadry2017towards}) and black-box methods (ZOO~\cite{chen2017zoo}).
	\item \textbf{VLM Attacks}. CADA~\cite{wang2025black} is utilized as a specialized adaptive attack tailored for AD systems to establish a worst-case performance standard. The evaluation also extends to threats targeting cross-modal alignment including  AdvCLIP~\cite{zhou2023advclip}, AnyAttack~\cite{zhang2024anyattack}, and SGA~\cite{lu2023set}, verifying resilience against modality-aware perturbations.
	\item \textbf{Patch Attacks}. This category involves localized threats created by superimposing adversarial patches generated via CADA and AdvCLIP onto physical traffic indicators.
\end{itemize}

\subsubsection{Compared Defenses}

Since there are barely any defense frameworks that simultaneously have detection, purification, and inference optimization capabilities, we compared their respective representative methods. The effectiveness of NutVLM is benchmarked against four distinct categories of defensive strategies spanning from input pre-processing to output rectification. 

\begin{itemize}
	\item \textbf{Image Transformation}. JPEG compression~\cite{guo2018countering}, median smoothing (MS)~\cite{chiang2020detection}, and Total Variation Minimization (TVM)~\cite{guo2018countering} are three widely-employed image transformation defense techniques used to disrupt adversarial perturbations before model input.
	\item \textbf{Image Denoising}. Neural Representation Purification (NRP)~\cite{naseer2020self} utilizes a pre-trained purifier network to reconstruct clean visual representations and eliminate adversarial noise before the inference stage to preserve feature integrity.
	\item \textbf{Adversarial Detection}.  Bit-depth compression technology~\cite{xu2018feature} identifies inputs with abnormal features by reducing the color bit precision of the input image, enabling the system to reject tampered samples or replace them with corresponding clean samples.
	\item \textbf{Output Post-processing}. AAA~\cite{chen2022adversarial} implements rule-based filtering for decision-level correction to detect extreme driving predictions and rectify them into safe control commands.
\end{itemize}

\begin{figure*}[t]
	\centering
	\begin{minipage}{0.48\textwidth}
		\centering
		\includegraphics[width=\textwidth]{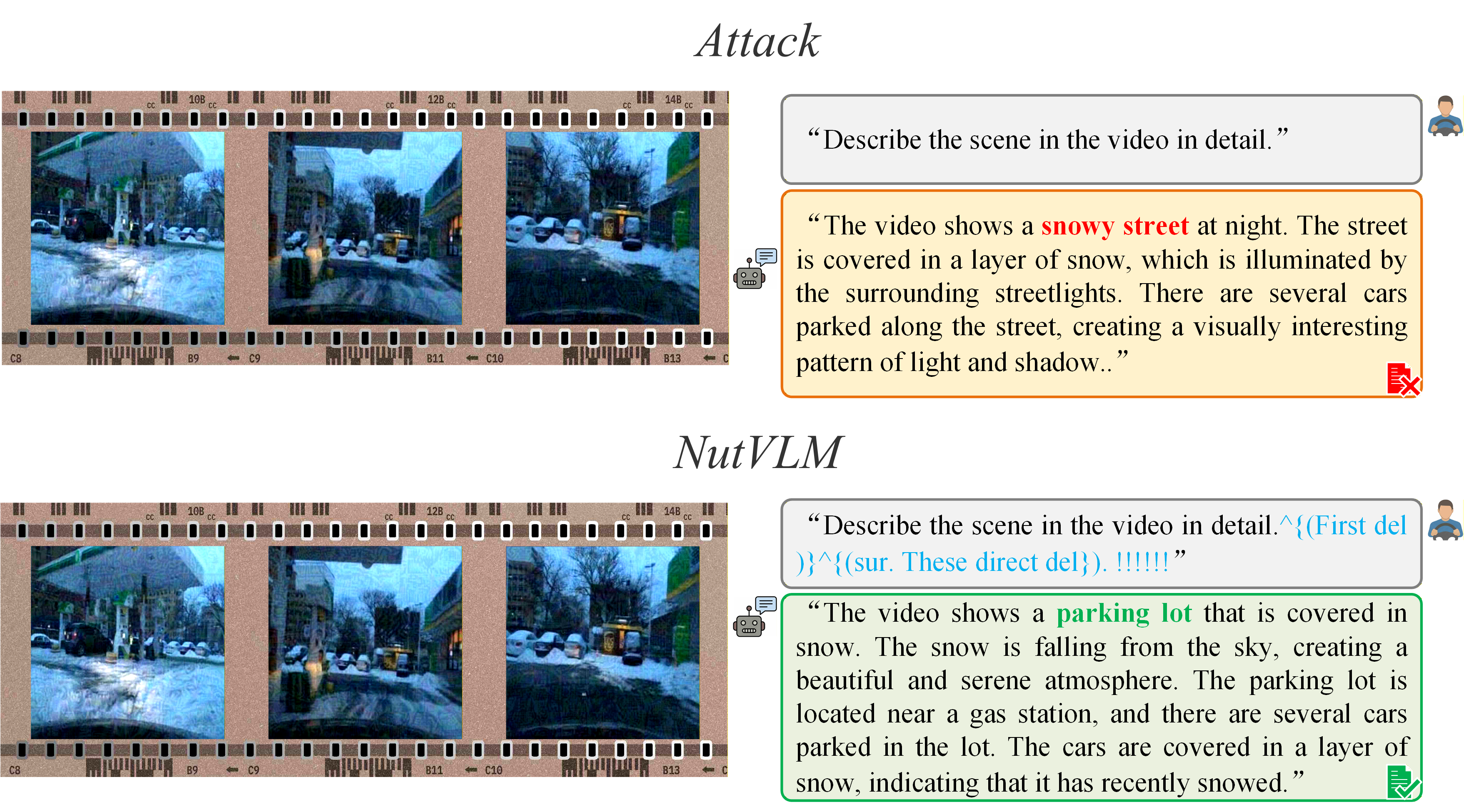}
		\makebox[\textwidth]{\footnotesize (\textbf{a}) CADA-Sense dataset}
		\label{fig_4_1}
	\end{minipage}
	\hspace{0.5cm}
	\begin{minipage}{0.48\textwidth}
		\centering
		\includegraphics[width=\textwidth]{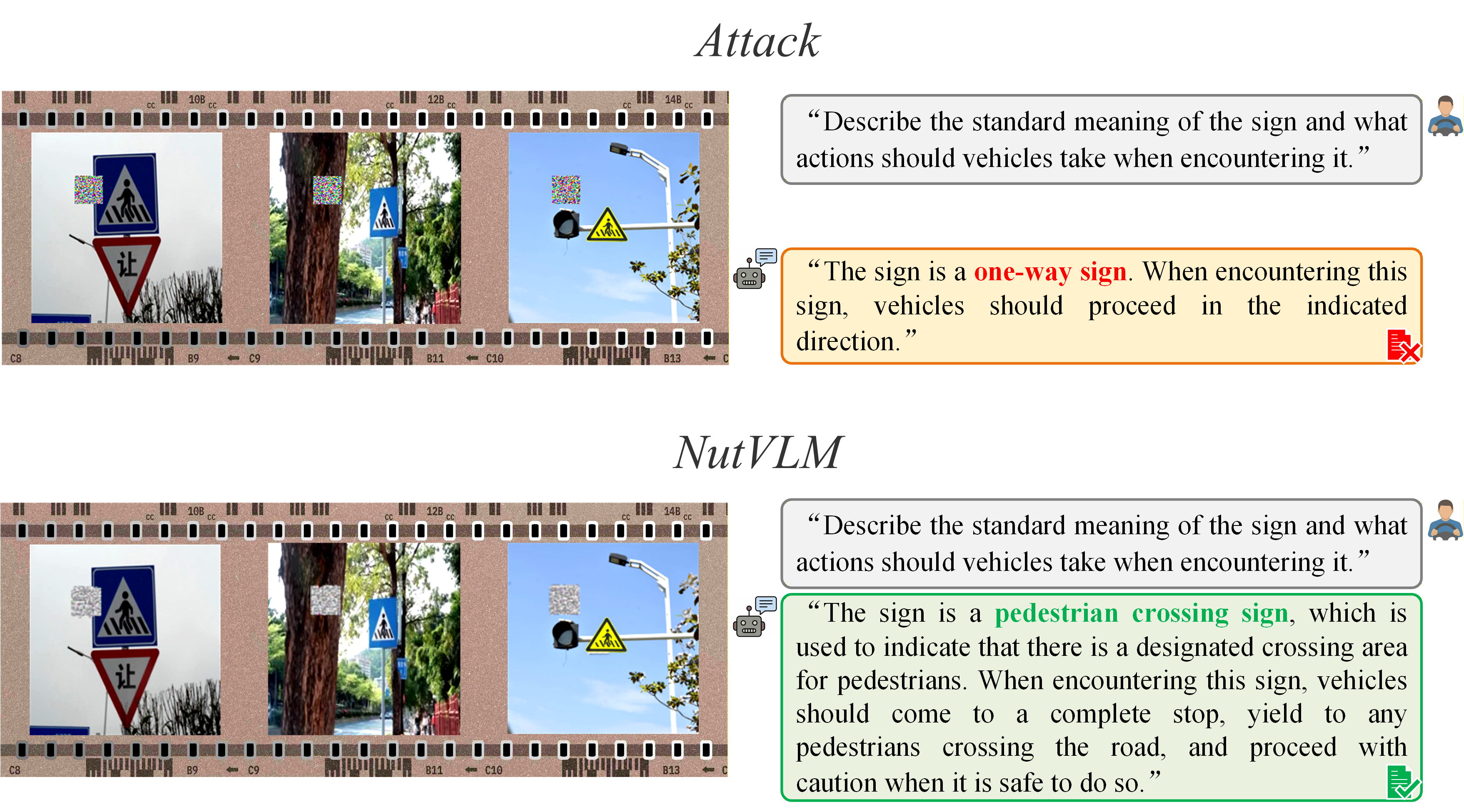}
		\makebox[\textwidth]{\footnotesize (\textbf{b}) CADA-Obj dataset}
		\label{fig_4_2}
	\end{minipage}
	\caption{Qualitative performance of NutVLM in safety-critical driving scenarios. (a) Scene description recovery on CADA-Sense, and (b) sign recognition on CADA-Obj. NutVLM effectively mitigates adversarial hallucinations (red) and restores accurate reasoning (green).}
	\label{fig_4}
\end{figure*}

\begin{table}[t]
	\centering
	\caption{Comparison of defense performance (Average Score) (\%) under different attack levels on the Dolphins. \textbf{Bold text} indicates the best performance in each level. \colorbox{gray!15}{\textbf{Gray cells}} indicate the results of our method.}
	\label{table_1}
	\footnotesize
	\renewcommand{\arraystretch}{1.25}
	\setlength{\tabcolsep}{2pt}
	\begin{tabularx}{\linewidth}{c l Y Y Y Y Y Y}
		\toprule
		\textbf{Lvl} & \textbf{Method} & \textbf{Weath.}\textcolor{blue}{$\uparrow$} & \textbf{Traf.}\textcolor{blue}{$\uparrow$} & \textbf{Time}\textcolor{blue}{$\uparrow$} & \textbf{Scene}\textcolor{blue}{$\uparrow$} & \textbf{Obj.}\textcolor{blue}{$\uparrow$} & \textbf{Desc.}\textcolor{blue}{$\uparrow$} \\
		\midrule
		\multirow{2}{*}{0} 
		& Scene-CADA & 47.75 & 62.20 & 36.79 & 47.45 & 37.04 & \textbf{38.96} \\
		& NutVLM 
		& \cellcolor{gray!15}\textbf{51.14} & \cellcolor{gray!15}\textbf{63.20} & \cellcolor{gray!15}\textbf{37.96} & \cellcolor{gray!15}\textbf{48.97} & \cellcolor{gray!15}\textbf{41.12} & \cellcolor{gray!15}38.61 \\
		\addlinespace[4pt]
		\multirow{2}{*}{1} 
		& Scene-CADA & 44.06 & 58.09 & 38.40 & \textbf{48.29} & \textbf{46.05} & \textbf{38.24} \\
		& NutVLM 
		& \cellcolor{gray!15}\textbf{46.93} & \cellcolor{gray!15}\textbf{60.16} & \cellcolor{gray!15}\textbf{41.16} & \cellcolor{gray!15}47.66 & \cellcolor{gray!15}43.15 & \cellcolor{gray!15}35.19 \\
		\addlinespace[4pt]
		\multirow{2}{*}{2} 
		& Scene-CADA & 40.79 & 58.00 & 37.77 & \textbf{47.45} & 40.00 & \textbf{39.12} \\
		& NutVLM 
		& \cellcolor{gray!15}\textbf{44.93} & \cellcolor{gray!15}\textbf{59.71} & \cellcolor{gray!15}\textbf{40.36} & \cellcolor{gray!15}47.34 & \cellcolor{gray!15}\textbf{43.22} & \cellcolor{gray!15}34.27 \\
		\addlinespace[4pt]
		\multirow{2}{*}{3} 
		& Scene-CADA & 38.16 & \textbf{63.70} & 38.02 & 46.73 & 38.24 & \textbf{36.50} \\
		& NutVLM 
		& \cellcolor{gray!15}\textbf{41.91} & \cellcolor{gray!15}63.33 & \cellcolor{gray!15}\textbf{41.65} & \cellcolor{gray!15}\textbf{46.76} & \cellcolor{gray!15}\textbf{40.99} & \cellcolor{gray!15}32.89 \\
		\addlinespace[4pt]
		\multirow{2}{*}{4} 
		& Scene-CADA & 35.40 & \textbf{59.03} & 40.47 & 46.69 & 33.82 & \textbf{36.44} \\
		& NutVLM 
		& \cellcolor{gray!15}\textbf{38.50} & \cellcolor{gray!15}54.77 & \cellcolor{gray!15}\textbf{45.28} & \cellcolor{gray!15}\textbf{47.03} & \cellcolor{gray!15}\textbf{35.84} & \cellcolor{gray!15}28.96 \\
		\bottomrule
	\end{tabularx}
\end{table}

\subsubsection{Evaluation Metrics}
To quantitatively evaluate the performance of NutVLM, we employ a multi-dimensional metric system spanning anomaly detection and linguistic task completion.

\begin{itemize} 
	\item \textbf{Detection Metrics}. NutNet++ is evaluated using F1-score, Detection Accuracy (D-Acc), Average Precision (AP), and Area Under the Curve (AUC) to assess its sensitivity to adversarial attacks. 
	\item \textbf{VLM Metrics}. Following standard protocols , we employ a multi-dimensional metric system for global attacks: Accuracy, Language Score (e.g., BLEU~\cite{papineni2002bleu}, METEOR~\cite{banerjee2005meteor}, and CIDEr~\cite{vedantam2015cider}), and GPT Score~\cite{liu2024geval}. For statistical efficiency in subsequent analysis, we define an \textit{Average Score} as the mean of these metrics. The \textit{Final Score} is then derived by averaging across all six driving tasks. Local attacks are evaluated exclusively via the \textit{GPT Score} using GPT-4o-mini to ensure consistent assessment. 
\end{itemize}

Across all metrics higher values (\textcolor{blue}{$\uparrow$}) signify improved defensive performance.

\begin{table}[t]
	\centering
	\caption{Comparison of defense performance (Final Score) (\%) among different defense methods across different attack levels on the Dolphins.}
	\label{table_2}
	\footnotesize
	\renewcommand{\arraystretch}{1.25}
	\setlength{\tabcolsep}{7.2pt} 
	\begin{tabular}{l c c c c c}
		\toprule
		\textbf{Method} & 
		\textbf{Lvl.~0}\textcolor{blue}{$\uparrow$} & 
		\textbf{Lvl.~1}\textcolor{blue}{$\uparrow$} & 
		\textbf{Lvl.~2}\textcolor{blue}{$\uparrow$} & 
		\textbf{Lvl.~3}\textcolor{blue}{$\uparrow$} & 
		\textbf{Lvl.~4}\textcolor{blue}{$\uparrow$} \\
		\midrule
		Scene--CADA & 45.03 & 45.52 & 43.58 & 43.56 & 40.86 \\
		\midrule
		JPEG~\cite{guo2018countering} & 43.20 & 42.85 & 41.50 & 41.10 & 39.25 \\
		TVM~\cite{guo2018countering}  & 38.15 & 37.60 & 36.45 & 35.80 & 34.20 \\
		MS~\cite{chiang2020detection}   & 45.25 & 44.55 & 43.75 & 43.70 & 40.05 \\
		\midrule
		Bit-Red~\cite{xu2018feature} & 45.10 & 45.59 & 44.10 & 43.20 & 40.60 \\
		\midrule
		NRP~\cite{naseer2020self} & 45.55 & 45.68 & 44.80 & 43.45 & 40.75 \\
		\midrule
		AAA~\cite{chen2022adversarial} & 45.05 & 45.64 & 44.45 & 43.95 & 40.45 \\
		\midrule
		NutVLM & \cellcolor{gray!15}\textbf{46.83} & \cellcolor{gray!15}\textbf{45.71} & \cellcolor{gray!15}\textbf{44.97} & \cellcolor{gray!15}\textbf{44.59} & \cellcolor{gray!15}\textbf{41.74} \\
		\bottomrule
	\end{tabular}
\end{table}

\subsection{Defense against Primary Attacks}

To evaluate the defense performance under worst-case scenarios, we employ the SOTA adaptive attack, CADA~\cite{wang2025black} as the primary threat, conducting extensive validation on the Dolphins. We assess NutVLM across diverse driving scenarios under varying attack intensities and compare its efficacy against several established defense paradigms.

\subsubsection{Analysis of Multi-task Robustness} The overall average of Accuracy, Language Score, and GPT Score across all tasks are utilized to evaluate defensive performance for a clearer presentation of the results. As detailed in Table~\ref{table_1}, NutVLM exhibits comprehensive resilience across six semantic understanding tasks under CADA attack. Notably, even in the absence of attacks (Level 0), our method provides a consistent performance, suggesting that the EAPT module further refines general feature representations. At the highest attack intensity, NutVLM achieves a remarkable recovery in the \textit{Weather} and \textit{Time} scenarios, outperforming the baseline by 3.10\% and 4.81\%, respectively. Cross-referencing these results with the Final Score in Table~\ref{table_2}, we observe that while all models exhibit performance declines as adversarial intensity increases, NutVLM consistently retains a superior performance margin and demonstrates greater resilience compared to standard VLMs. This indicates that our framework effectively preserves a higher degree of critical semantic information even under severe global perturbations. Qualitatively, the scene description recovery visualized in Fig.~4(a) demonstrates NutVLM's ability to suppress adversarial hallucinations and restore semantic coherence in complex driving environments.

\subsubsection{Comparison with Other Defenses} 
Table~\ref{table_2} benchmarks NutVLM against established defense mechanisms, where it consistently achieves the highest Final Score across all attack levels. The results demonstrate that isolated improvements in a single defensive dimension are insufficient for robust protection because competing paradigms lose their efficacy as attack intensity escalates. Specifically, NutVLM outperforms the representative methods NRP, AAA, along with Bit-Red, by $0.72\%$, $0.86\%$ as well as $1.05\%$ in the comprehensive score, respectively. These improvements validate the structural advantage of the framework, which synergistically integrates visual restoration with semantic correction to surpass traditional defenses relying solely on image-level heuristics.

\begin{table}[t]
	\centering
	\caption{Comparison of defense performance (GPT Score) (\%) between Obj-CADA and NutVLM across different attack levels on the Dolphins.}
	\label{table_3}
	\footnotesize
	\renewcommand{\arraystretch}{1.25}
	\setlength{\tabcolsep}{2pt}
	\begin{tabularx}{\linewidth}{l Y Y Y Y Y Y}
		\toprule
		\textbf{Method} & 
		\textbf{Lvl.~0}\textcolor{blue}{$\uparrow$} & 
		\textbf{Lvl.~1}\textcolor{blue}{$\uparrow$} & 
		\textbf{Lvl.~2}\textcolor{blue}{$\uparrow$} & 
		\textbf{Lvl.~3}\textcolor{blue}{$\uparrow$} & 
		\textbf{Lvl.~4}\textcolor{blue}{$\uparrow$} & 
		\textbf{Avg.}\textcolor{blue}{$\uparrow$} \\ 
		\midrule
		Obj-CADA & 55.57 & 52.46 & 52.36 & 51.14 & 50.86 & 52.48 \\
		\addlinespace[4pt]
		NutVLM &\cellcolor{gray!15}\textbf{56.07} &\cellcolor{gray!15}\textbf{58.82} &\cellcolor{gray!15}\textbf{56.86} &\cellcolor{gray!15}\textbf{56.14} &\cellcolor{gray!15}\textbf{53.64} &\cellcolor{gray!15}\textbf{56.31} \\
		\bottomrule
	\end{tabularx}
\end{table}

\begin{table}[t]
	\centering
	\caption{Comparison of defense performance (Final Score) (\%) with different VLM backbones under various attacks.}
	\label{table_5}
	\footnotesize
	\renewcommand{\arraystretch}{1.25}
	\setlength{\tabcolsep}{4pt}
	\begin{tabularx}{\linewidth}{l Y Y Y Y}
		\toprule
		\textbf{Method} & \textbf{DP}~\cite{ma2024dolphins}\textcolor{blue}{$\uparrow$} & \textbf{IB}~\cite{dai2023instructblip}\textcolor{blue}{$\uparrow$} & \textbf{LV}~\cite{liu2023visual}\textcolor{blue}{$\uparrow$} & \textbf{MG}~\cite{zhu2023minigpt}\textcolor{blue}{$\uparrow$} \\
		\midrule
		Org & 45.03 & 38.19 & 43.93 & 26.89 \\
		\midrule
		
		FGSM~\cite{goodfellow2014explaining} & \textbf{44.43} & 35.36 & 40.89 & 28.32 \\
		NutVLM & \cellcolor{gray!15}43.16 & \cellcolor{gray!15}\textbf{35.99} & \cellcolor{gray!15}\textbf{45.70} & \cellcolor{gray!15}\textbf{31.19} \\
		\addlinespace[4pt]
		
		PGD~\cite{mkadry2017towards} & 40.27 & 36.28 & 43.91 & 27.61 \\
		NutVLM & \cellcolor{gray!15}\textbf{45.90} & \cellcolor{gray!15}\textbf{38.75} & \cellcolor{gray!15}\textbf{48.28} & \cellcolor{gray!15}\textbf{30.07} \\
		\addlinespace[4pt]
		
		ZOO~\cite{chen2017zoo} & 45.74 & 36.38 & 43.61 & 25.35 \\
		NutVLM & \cellcolor{gray!15}\textbf{47.20} & \cellcolor{gray!15}\textbf{38.01} & \cellcolor{gray!15}\textbf{50.58} & \cellcolor{gray!15}\textbf{28.68} \\
		\addlinespace[4pt]
		
		AdvClip~\cite{zhou2023advclip} & 47.44 & \textbf{37.47} & 42.79 & 26.48 \\
		NutVLM & \cellcolor{gray!15}\textbf{49.10} & \cellcolor{gray!15}37.24 & \cellcolor{gray!15}\textbf{49.27} & \cellcolor{gray!15}\textbf{31.04} \\
		\addlinespace[4pt]
		
		AnyAttack~\cite{zhang2024anyattack} & 43.45 & 33.74 & 42.54 & 23.71 \\
		NutVLM & \cellcolor{gray!15}\textbf{45.31} & \cellcolor{gray!15}\textbf{35.97} & \cellcolor{gray!15}\textbf{46.34} & \cellcolor{gray!15}\textbf{27.92} \\
		\addlinespace[4pt]
		
		SGA~\cite{lu2023set} & 46.42 & 37.84 & 44.30 & 28.32 \\
		NutVLM & \cellcolor{gray!15}\textbf{49.87} & \cellcolor{gray!15}\textbf{38.80} & \cellcolor{gray!15}\textbf{50.46} & \cellcolor{gray!15}\textbf{31.12} \\
		\bottomrule
	\end{tabularx}
\end{table}

\subsubsection{Defense against Patch Attacks} 
A key highlight of NutVLM is its robustness against local threats, which is further substantiated through patch attack evaluations in Table~\ref{table_3}. NutVLM exhibits exceptional stability by maintaining an average GPT Score of 56.31\% to achieve a 3.83\% improvement over the baseline performance of 52.48\%. Even under the most challenging Level 4 attacks, the system sustains a score of 53.64\%, representing a marginal decline of only 1.93\% compared to the clean sample performance of 55.57\%. Furthermore, as AdvClip also utilizes patch-based perturbations, Table~\ref{table_5} corroborates that NutVLM provides broad and substantial defensive gains across diverse backbones. While the defensive margin slightly diminishes as adversarial patches increase in size due to the occlusion of essential visual cues, NutVLM still effectively mitigates adversarial semantic distortions. This is qualitatively supported by Fig.~4(b), where NutVLM successfully identifies obscured traffic signs by neutralizing local adversarial patches to ensure safe driving logic.

In summary, NutVLM represents a significant leap in adversarial defense for AD VLMs, delivering robust and consistent protection across multi-granular tasks and varying attack intensities.

\subsection{Defense Generalization Capabilities}

To assess the generalization and reliability of NutVLM, we conduct evaluation from two critical perspectives: transferability across diverse VLM backbones and resilience against various adversarial attacks.

\subsubsection{Stability under Different Backbones} 
NutVLM maintains a remarkably stable performance profile across different model architectures as adversarial pressure increases from Level 0 to Level 4. According to the results in Table~\ref{table_4}, our method consistently outperforms the Scene-CADA baseline on the IB, LV, and MG. At the most severe intensity (Level 4), NutVLM preserves a Final Score of 1.95\% on IB and 2.14\% on LV over the respective baseline scores. While the defensive improvement on the MG backbone is less pronounced, the framework still demonstrates a performance increase, which is likely influenced by the inability to directly access the language encoder of MG. These observations confirm that the semantic regularization effectively prevents catastrophic performance collapse across various VLM architectures.

\begin{table}[t]
	\centering
	\caption{Comparison of defense performance (Final Score) (\%) with different VLM backbones under different attack levels.}
	\label{table_4}
	\footnotesize
	\renewcommand{\arraystretch}{1.25}
	\setlength{\tabcolsep}{2pt}
	\begin{tabularx}{\linewidth}{c l Y Y Y Y Y}
		\toprule
		\textbf{Metric} & \textbf{Method} & \textbf{Lvl.~0}\textcolor{blue}{$\uparrow$}  & \textbf{Lvl.~1}\textcolor{blue}{$\uparrow$}  & \textbf{Lvl.~2}\textcolor{blue}{$\uparrow$}  & \textbf{Lvl.~3}\textcolor{blue}{$\uparrow$}  & \textbf{Lvl.~4}\textcolor{blue}{$\uparrow$}  \\
		\midrule
		\multirow{2}{*}{IB~\cite{dai2023instructblip}} 
		& Scene--CADA & 36.08 & 34.74 & 33.72 & 32.69 & 30.60 \\
		& NutVLM 
		& \cellcolor{gray!15}\textbf{37.19} & \cellcolor{gray!15}\textbf{36.44} & \cellcolor{gray!15}\textbf{35.89} & \cellcolor{gray!15}\textbf{33.03} & \cellcolor{gray!15}\textbf{32.55} \\
		\addlinespace[4pt]
		\multirow{2}{*}{LV~\cite{liu2023visual}} 
		& Scene--CADA & 44.12 & 44.81 & 41.47 & 37.83 & 38.33 \\
		& NutVLM 
		& \cellcolor{gray!15}\textbf{44.54} & \cellcolor{gray!15}\textbf{45.34} & \cellcolor{gray!15}\textbf{42.03} & \cellcolor{gray!15}\textbf{39.99} & \cellcolor{gray!15}\textbf{40.47} \\
		\addlinespace[4pt]
		\multirow{2}{*}{MG~\cite{zhu2023minigpt}} 
		& Scene--CADA & 26.03 & 27.38 & 23.63 & 24.23 & 24.38 \\
		& NutVLM 
		& \cellcolor{gray!15}\textbf{28.00} & \cellcolor{gray!15}\textbf{27.86} & \cellcolor{gray!15}\textbf{27.29} & \cellcolor{gray!15}\textbf{24.78} & \cellcolor{gray!15}\textbf{24.52} \\
		\bottomrule
	\end{tabularx}
\end{table}

\subsubsection{Resilience against Diverse Attacks} 
As shown in Table~\ref{table_5}, our framework consistently achieves superior defense performance, neutralizing diverse perturbation patterns across almost all Final Scores. NutVLM exhibits exceptional resilience against the six representative attacks introduced in the aforementioned sections. With the LLaVA backbone, NutVLM provides substantial gains and often restores performance to levels that exceed the original unattacked baseline (Org). For instance, under the SGA attack, NutVLM achieves a Final Score of 50.46\%, reflecting a 6.53\% performance gain over the baseline of 43.93\%. This consistent dominance across various attacks demonstrates that the semantic alignment enforced by our EAPT module is highly agnostic to the specific optimization algorithms employed by different attackers.

\begin{figure*}[t]
	\centering
	\begin{minipage}{0.475\textwidth}
		\centering
		\includegraphics[width=\textwidth]{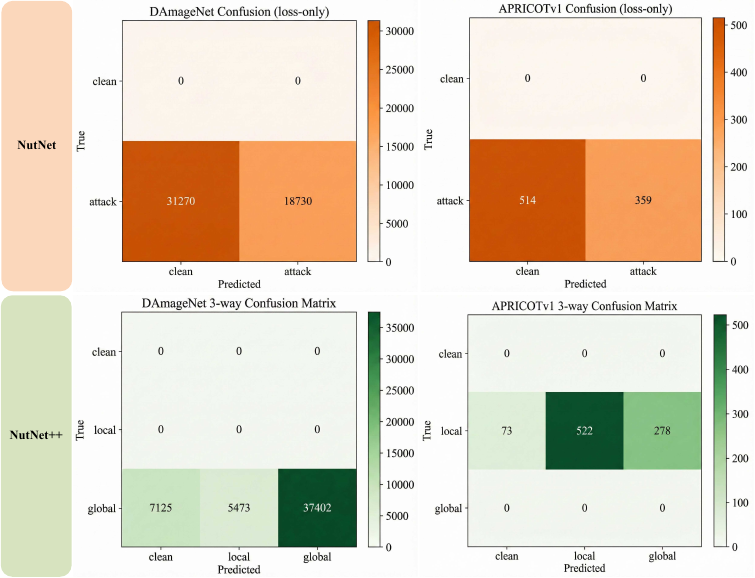}
		\makebox[\textwidth]{\footnotesize (\textbf{a}) General dataset}
		\label{fig_5_1}
	\end{minipage}
	\hspace{0.5cm}
	\begin{minipage}{0.48\textwidth}
		\centering
		\includegraphics[width=\textwidth]{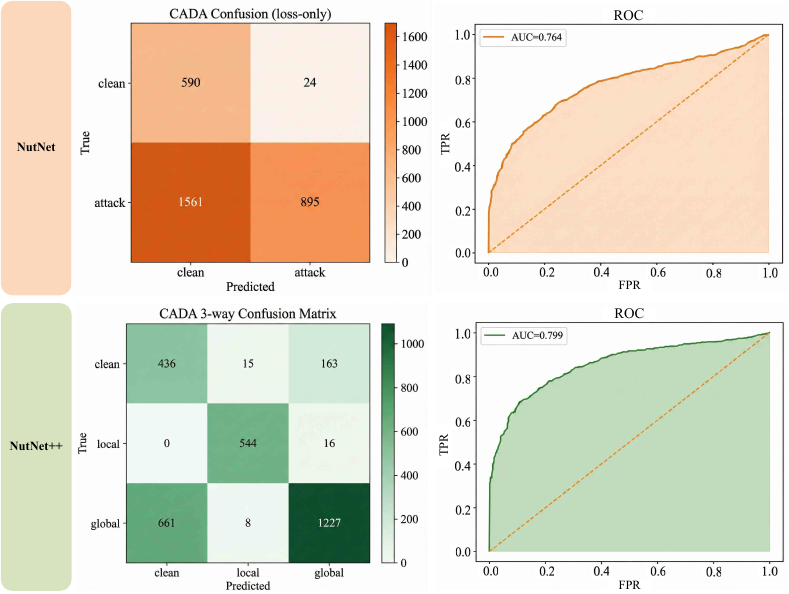}
		\makebox[\textwidth]{\footnotesize (\textbf{b}) AD-specific datasets}
		\label{fig_5_2}
	\end{minipage}
	\caption{Quantitative evaluation of NutNet++ performance. (\textbf{a}) Confusion matrices on general datasets (DAmageNet and APRICOT), and (\textbf{b}) Confusion matrices and ROC curves on AD-specific datasets (CADA).}
	\label{fig_5}
\end{figure*}

Taken together, these findings underscore the robust generalization of NutVLM, positioning it as a versatile defense framework that remains effective across heterogeneous VLM backbones and diverse adversarial conditions.

\begin{table}[t]
	\centering
	\caption{Performance comparison (\%) of detection efficacy on general datasets between NutNet and NutNet++. \textbf{Bold text} indicates the best performance for each metric. \colorbox{gray!15}{\textbf{Gray cells}} specifically represent the binary classification results of our NutNet++ method.}
	\label{table_6}
	\footnotesize 
	\renewcommand{\arraystretch}{1.25}
	\setlength{\tabcolsep}{2pt} 
	\begin{tabularx}{\linewidth}{l Y Y Y Y}
		\toprule
		\multirow{2}{*}{\textbf{Datasets}} & \multicolumn{2}{c}{\textbf{DAmageNet (Global)~\cite{chen2020universal}}} & \multicolumn{2}{c}{\textbf{APRICOT (Local)~\cite{braunegg2020apricot}}} \\
		\cmidrule(lr){2-3} \cmidrule(lr){4-5}
		& \textbf{F1}\textcolor{blue}{$\uparrow$} & \textbf{D-Acc}\textcolor{blue}{$\uparrow$} & \textbf{F1}\textcolor{blue}{$\uparrow$} & \textbf{D-Acc}\textcolor{blue}{$\uparrow$} \\
		\midrule
		NutNet & 54.50 & 37.46 & 58.28 & 41.12 \\
		\addlinespace[4pt] 
		& \textbf{85.59} & \textbf{74.80} & \textbf{74.80} & \textbf{59.80} \\
		\multirow{-2}{*}{NutNet++} 
		& \cellcolor{gray!15}\textbf{92.33} 
		& \cellcolor{gray!15}\textbf{85.75} 
		& \cellcolor{gray!15}\textbf{92.25}
		& \cellcolor{gray!15}\textbf{85.60} \\
		\bottomrule
	\end{tabularx}
\end{table}

\subsection{Ablation Studies}
This section presents a series of ablation experiments to substantiate the detection precision of NutNet++ and the architectural effectiveness of the EAPT module along with the synergistic efficacy of the dual-branch defense framework.

\subsubsection{Efficacy of NutNet++ Detection}

To verify the robustness and generalization of NutNet++, we evaluated its performance on two types of benchmarks.

As presented in Fig.~\ref{fig_5} and Table~\ref{table_6}, NutNet++ demonstrates substantial improvements over the original NutNet baseline. Quantitatively, on the DAmageNet dataset, our method boosts the F1-score and D-Acc by margins of 31.09\% (37.83\%) and 37.34\% (48.29\%), respectively. Consistent gains are observed on the APRICOT dataset, with F1-score and D-Acc increasing by 16.52\% (33.97\%) and 18.68\% (44.48\%). Notably, values in parentheses denote the performance for binary (Attack vs. Clean) classification, highlighting the efficacy and robustness of the proposed Soft-Recall Mechanism. Confusion matrices in Fig.~\ref{fig_5}(a) illustrate a significant reduction in misclassification errors while spatial visualizations in Fig.~\ref{fig_4}(b) confirm precise patch localization and masking in physical scenes. These findings substantiate the general detection and purification capabilities of NutNet++.

\begin{table}[t]
	\centering
	\caption{Performance comparison (\%) of detection efficacy on AD-specific datasets between NutNet and NutNet++.}
	\label{table_7}
	\footnotesize
	\renewcommand{\arraystretch}{1.25}
	\setlength{\tabcolsep}{4pt}
	\begin{tabularx}{\linewidth}{l Y Y Y Y}
		\toprule
		\textbf{Method} & \textbf{F1}\textcolor{blue}{$\uparrow$} & \textbf{D-Acc}\textcolor{blue}{$\uparrow$} & \textbf{AP}\textcolor{blue}{$\uparrow$} & \textbf{AUC}\textcolor{blue}{$\uparrow$} \\
		\midrule
		NutNet & 53.04 & 48.37 & 93.40 & 76.40 \\
		NutNet++&\cellcolor{gray!15}\textbf{73.94}&\cellcolor{gray!15}\textbf{71.89}&\cellcolor{gray!15}\textbf{94.40}&\cellcolor{gray!15}\textbf{79.90} \\
		\bottomrule
	\end{tabularx}
\end{table}

Moving beyond general benchmarks, we assessed the model's defensive capability in safety-critical driving scenarios using the CADA dataset across levels 0 to 4. As indicated in Table~\ref{table_7}, NutNet++ achieves a substantial improvement in F1-score (73.94\%) and D-Acc (71.89\%) compared to the baseline values of 53.04\% and 48.37\% for NutNet. The superior AUC of 79.90\% and the ROC curves in Fig.~\ref{fig_5}(b) further validate its role as a robust sentinel that effectively extends the defensive boundary of the VLMs. While the confusion matrices show certain false detection rates, this is primarily caused by weak adversarial intensity at Level 1. According to Table~\ref{table_2}, such subtle perturbations occasionally function as data augmentation, which makes them difficult to distinguish from clean samples. Qualitatively, as shown in Fig.~\ref{fig_4}(b), NutNet++ enables the model to provide accurate semantic answers and correct reasoning even when the visual input is compromised by adversarial patches, effectively neutralizing the "hallucinations" that lead to dangerous driving decisions.

\subsubsection{Design Rationale of the EAPT Module}
\begin{figure}[t]
	\centering	
	{\includegraphics[scale=0.28]{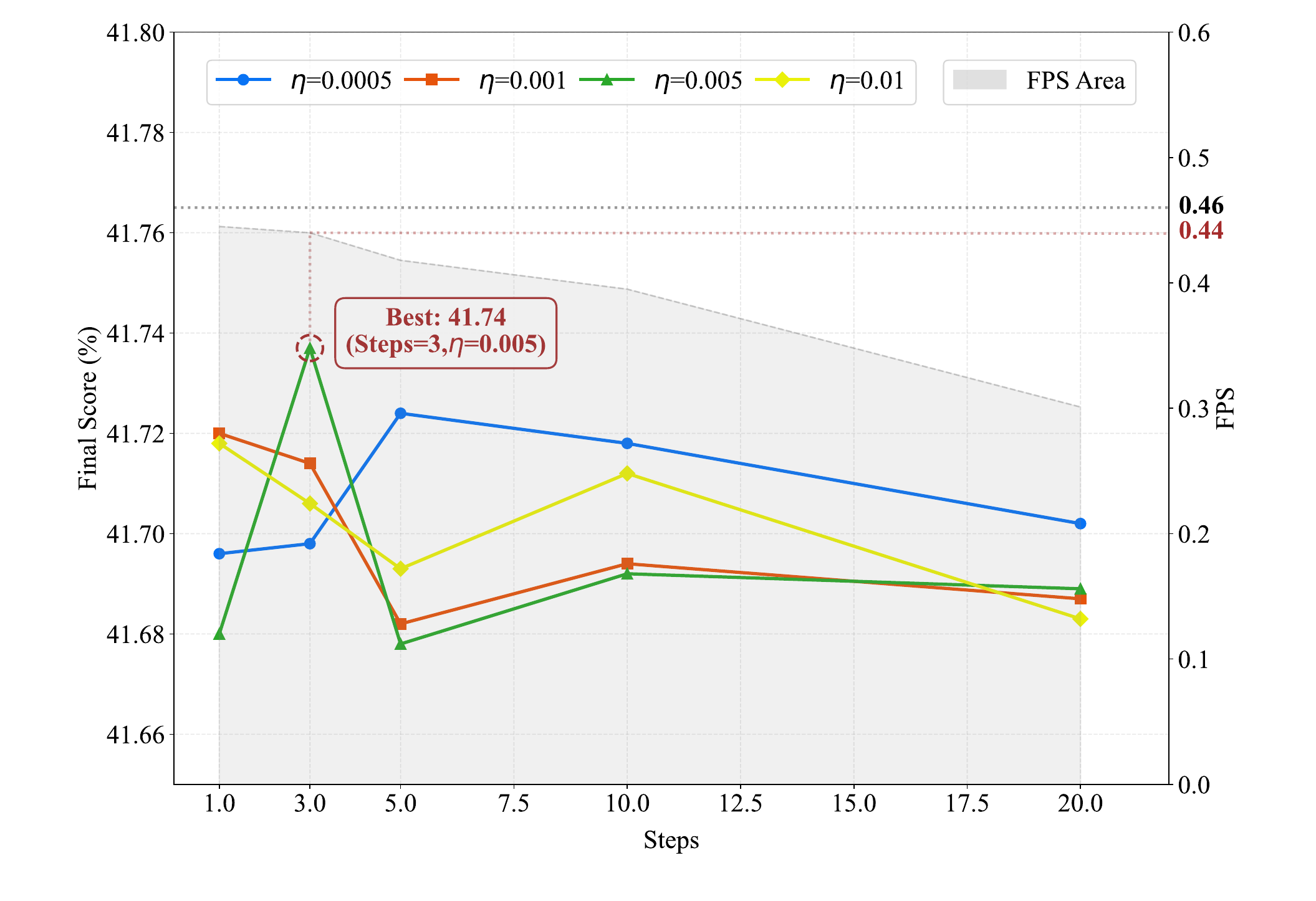}}
	\captionsetup{font={footnotesize}}
	\caption{Performance analysis (Final Score) of the EAPT module under varying steps and learning rates.}
	\label{fig_6}
\end{figure}

Since real-time processing is paramount for AD tasks, we introduce Frames Per Second (FPS) as a critical evaluation metric for EAPT to balance effectiveness and efficiency. We conducted an ablation study on two pivotal hyperparameters: the number of optimization steps and the learning rate ($\eta$). As illustrated in the integrated analysis in Fig.~\ref{fig_6}, the Final Score generally declines as the number of steps increases beyond a certain threshold, suggesting that excessive iterations may lead to overfitting or the introduction of secondary semantic distortions. Simultaneously, the FPS (represented by the gray shaded area) decreases as the step count increases, confirming the computational overhead of prolonged refinement. By navigating this trade-off, we identify an optimal configuration (Steps=3, $\eta=0.005$) that achieves a peak score of 41.74\% while maintaining a high throughput of 0.44 FPS. By comparison, the baseline FPS without the EAPT module is 0.46, indicating that our semantic correction introduces negligible latency. This optimized design enables the VLM to restore semantic integrity efficiently, demonstrating that rapid and precise correction is essential for deploying robust defense frameworks in time-sensitive AD environments.

\subsubsection{Decoupling Analysis of the Dual-Branch Defense}
We perform a decoupling analysis to examine the synergy between visual restoration and semantic alignment branches by evaluating three specific configurations.
\begin{itemize}
	\item \textbf{NutNet++\_Mask} applies pixel-level grayscale occlusion to neutralize identified adversarial regions within the visual input.
	\item \textbf{NutNet++\_Patch} utilizes defensive alert signs by either placing them in the image corner for global attacks or directly covering the adversarial patch for local attacks.
	\item \textbf{NutNet++\_EAPT} implements instruction correction through our proposed EAPT module to ensure cross-modal semantic consistency.
\end{itemize}

To investigate the individual contribution of each defense component, we conduct a 'decoupling analysis' by isolating visual and semantic branches. Experimental results demonstrate that the NutNet++\_Patch variant exhibits suboptimal performance across both global and local scenarios as evidenced by the lower average scores of 44.43\% in Table~\ref{table_8} and 49.18\% in Table~\ref{table_9} relative to other configurations within the same tables. This degradation occurs because the inclusion of alert markers within the visual input introduces extraneous semantic noise that distracts the VLM. Consequently, such indicators are more effectively implemented within the user interface rather than being processed as direct model inputs. Regarding global digital perturbations in Table~\ref{table_8}, the NutNet++\_Mask configuration (44.80\%) proves less effective because pixel-level occlusion across the entire scene inadvertently obscures critical contextual information. In contrast, NutNet++\_EAPT achieves the highest performance under global attacks (44.99\%) by utilizing semantic anchors to rectify distortions without sacrificing background integrity. However, as for local attacks in Table~\ref{table_9}, EAPT (54.01\%) underperforms relative to NutNet++\_Mask (56.31\%) because purely semantic soft constraints struggle to neutralize the hard interference of high-saliency patches. Unlike linguistic anchors, spatial masking remains more effective for local threats by physically eliminating the noise source at the visual feature level to provide a more robust defense. Overall, the strategic separation of visual masking for local threats and semantic correction for global perturbations ensures that NutVLM provides a comprehensive and context-aware defense against diverse adversarial strategies.
\begin{table}[t]
	\centering
	\caption{Performance comparison (Final Score) (\%) of decoupling strategies under global attacks. \colorbox{gray!15}{\textbf{Gray cells}} denote the configuration adopted in our NutVLM. Aver denotes the average Final Score across all attack levels.}
	\label{table_8}
	\footnotesize
	\renewcommand{\arraystretch}{1.25}
	\setlength{\tabcolsep}{3.5pt}
	\begin{tabularx}{\linewidth}{l Y Y Y Y Y Y}
		\toprule
		\textbf{Method} & \textbf{Lvl. 0}\textcolor{blue}{$\uparrow$} & \textbf{Lvl. 1}\textcolor{blue}{$\uparrow$} & \textbf{Lvl. 2}\textcolor{blue}{$\uparrow$} & \textbf{Lvl. 3}\textcolor{blue}{$\uparrow$} & \textbf{Lvl. 4}\textcolor{blue}{$\uparrow$} & \textbf{Aver}\textcolor{blue}{$\uparrow$} \\
		\midrule
		NutNet++\_Mask  & 45.99 & 46.36 & 44.91 & 44.36 & \textbf{42.37} & 44.80 \\
		NutNet++\_Patch & \textbf{47.15} & 45.85 & 44.44 & 43.07 & 41.62 & 44.43 \\
		NutNet++\_EAPT  & \cellcolor{gray!15}46.83 & \cellcolor{gray!15}\textbf{46.83} & \cellcolor{gray!15}\textbf{44.97} & \cellcolor{gray!15}\textbf{44.59} & \cellcolor{gray!15}41.74 & \cellcolor{gray!15}\textbf{44.99} \\
		\bottomrule
	\end{tabularx}
\end{table}
\begin{table}[t]
	\centering
	\caption{Performance comparison (Final Score) (\%) of decoupling strategies under local attacks.}
	\label{table_9}
	\footnotesize
	\renewcommand{\arraystretch}{1.25}
	\setlength{\tabcolsep}{3.5pt}
	\begin{tabularx}{\linewidth}{l Y Y Y Y Y Y}
		\toprule
		\textbf{Method} & \textbf{Lvl. 0}\textcolor{blue}{$\uparrow$} & \textbf{Lvl. 1}\textcolor{blue}{$\uparrow$} & \textbf{Lvl. 2}\textcolor{blue}{$\uparrow$} & \textbf{Lvl. 3}\textcolor{blue}{$\uparrow$} & \textbf{Lvl. 4}\textcolor{blue}{$\uparrow$} & \textbf{Aver}\textcolor{blue}{$\uparrow$} \\
		\midrule
		NutNet++\_Mask  & \cellcolor{gray!15}\textbf{56.07} & \cellcolor{gray!15}\textbf{58.82} & \cellcolor{gray!15}\textbf{56.86} & \cellcolor{gray!15}\textbf{56.14} & \cellcolor{gray!15}\textbf{53.64} & \cellcolor{gray!15}\textbf{56.31} \\
		NutNet++\_Patch & 51.96 & 50.18 & 48.18 & 48.11 & 47.46 & 49.18 \\
		NutNet++\_EAPT  & 54.89 & 55.11 & 55.11 & 52.89 & 52.07 & 54.01 \\
		\bottomrule
	\end{tabularx}
\end{table}	
\section{Conclusion}

This paper presents NutVLM, a self-adaptive defense framework designed to secure AD VLMs against full-dimension attacks through an integrated detection and purification strategy. NutNet++ provides three-way threat sensing and visual purification while EAPT executes real-time instruction correction through latent space optimization. Evaluations show that NutNet++ achieves a 73.94\% detection F1-score. Under challenging CADA Level 4 scenarios, NutVLM maintains a 41.74\% Final Score for global attacks and yields a 53.64\% GPT Score against local threats. The framework also generalizes effectively across diverse backbones including InstructBlip, LLaVA, and MiniGPT-v4. By attaining high inference speed, NutVLM provides the critical defense efficiency essential for safe navigation in AD systems. This approach ensures strong adversarial resilience and maintains clean-sample performance without full-model retraining. 

Future research will extend this framework to extreme environmental conditions and end-to-end planning loops. Additionally, integrating multi-sensory contexts like audio-visual fusion will further reinforce the long-term safety of intelligent transportation systems.

\bibliographystyle{IEEEtran}
\bibliography{mybibfile}

\vfill

\end{document}